\documentclass[copyright,creativecommons]{eptcs}
\usepackage{xcolor}
\usepackage{graphicx}
\usepackage{amsmath}
\usepackage{placeins}

\providecommand{\event}{AREA 2023} 

\usepackage{iftex}

\ifpdf
  \usepackage{underscore}         
  \usepackage[T1]{fontenc}        
\else
  \usepackage{breakurl}           
\fi

\title{Adaptive Application Behaviour for Robot Swarms using Mixed-Criticality}
\author{
Sven Signer
\institute{Department of Computer Science\\
University of York\\
York, United Kingdom}
\email{sven.signer@york.ac.uk}
\and
Ian Gray
\institute{Department of Computer Science\\
University of York\\
York, United Kingdom}
\email{ian.gray@york.ac.uk}
}
\def\titlerunning{Adaptive Application Behaviour for Robot Swarms using Mixed-Criticality}
\def\authorrunning{S. Signer \& I. Gray}
\begin{document}
\maketitle

\begin{abstract}
Communication is a vital component for all swarm robotics applications, and even simple swarm robotics behaviours often break down when this communication is unreliable. Since wireless communications are inherently subject to interference and signal degradation, real-world swarm robotics applications will need to be able handle such scenarios. This paper argues for tighter integration of application level and network layer behaviour, so that the application can alter its behaviour in response to a degraded network. This is systematised through the implementation of a mixed-criticality system model. We compare a static application behaviour with that of an application that is able to alter its behaviour in response to the current criticality level of a mixed-criticality wireless protocol. Using simulation results we show that while a static approach is sufficient if the network conditions are known a priori, a mixed-criticality system model is able to adapt application behaviour to better support unseen or unpredictable conditions.
\end{abstract}

\section{Introduction}

Swarm robotics platforms usually rely on wireless communications, which by their nature can be unreliable and exhibit unpredictable timing characteristics. The standard approach is to use best-effort protocols with sufficient redundancy so that all important traffic can be conveyed. Conventional protocols such as WiFi, ZigBee, and Bluetooth offer good throughput and robustness, but they rely on features like random backoff and retransmissions which can be disqualifying when attempting to build high-criticality systems that rely on timing correctness. There are similarly a range of swarm-oriented protocols such as Glossy~\cite{5779066}, and the related Low-Power Wireless Bus~\cite{10.1145/2426656.2426658} and Blink~\cite{10.1145/3012005} which all remain topology-agnostic by flooding the network with packets and retransmissions in order to maximise connectivity at the expense of throughput and power-efficiency. 

Accordingly there has been a more recent trend towards protocols which are timing-aware, such as WirelessHART~\cite{international2016industrial} and AirTight~\cite{airtight}. These approaches require a priori information about the system, but compensate by allowing for greater timing confidence. They still, however, suffer from the inherent unreliability of wireless links. 

This work combines ideas from these timing-aware protocols and from the real-time systems domain to argue for application-level adaptive behaviour to support reliability in the presence of errors and unreliable communications. If a swarm robotics platform employs a wireless protocol that can provide feedback as to whether communication reliability requirements are currently being met, the application can be designed to adapt its own behaviour in order to recover connectivity if packet delivery rates fall below a predefined threshold. This paper argues for the use of a mixed-criticality approach, in which the application switches between two behaviour modes depending on the network conditions.

\section{Application Adaptivity and Mixed Criticality Systems}

A common claim is that robot swarms achieve fault tolerance and redundancy through the size of the swarm. However as is well-understood in the research, communication failures are particularly problematic and can actually be compounded by increasing swarm size~\cite{bjerknes2013fault}. Swarm applications must be designed with the reliability of communications as a priority. Robotic swarm programming models like Buzz~\cite{pinciroli2016buzz} allow for developers to discuss higher level concepts such as moving groups of robots as one, but frameworks which tie reliability requirements from the application developer to the reliability state of the running system are less formalised.

This work argues that in order to address this issue, a swarm robotics system with unreliable communications should be viewed as a Mixed Criticality System (MCS), and that such a characterisation can be used to explicitly guide the system's behaviour at times of overload or error. The MCS model is a commonly-deployed system model in the domains of high-integrity or safety-critical systems, and is used to provide guarantees on the behaviour and timing of the system in the presence of unreliability. Informally, the developer provides a description of their system, along with an ``importance'' for each task, and the guarantees that are required to support these tasks at each criticality level. If the system enters a state when it can no longer guarantee the ``important'' parts, then a graceful degradation is codified into the model.

The original formulation of MCS~\cite{Vestal2007} was aimed at CPU scheduling problems and defined the system to have two criticality levels: \textit{LO} and \textit{HI}. Each task/process/job is designated either \textit{LO} criticality or \textit{HI} criticality: \textit{HI} tasks perform safety-critical functions and are required for safety assurance, whereas \textit{LO} tasks are less important and may occasionally fail or miss their deadlines. The response time analysis of CPU scheduling requires tasks to be assigned an estimate of their worst-case execution time (WCET) a priori, but determining this value proves difficult in practice. If a task's WCET comes from estimation or measurement it may be optimistic (i.e. not a true worst-case time). This could result in important tasks missing their deadlines if the true WCET at runtime is larger than the assumed value.  A task's WCET may also come from a safe, analytical approach which is guaranteed to cover the worst-case but in practice this might be very pessimistic. This could lead to the application designer falsely believing that the system is not schedulable without greater hardware resources, which may in turn be infeasible for other reasons (cost, energy use, etc). 

The key insight of the MCS model is that the optimistic execution times will be correct most of the time, and the true worst-case is usually only seen in rare situations. Tasks are therefore assigned two estimates of their worst-case execution time, a \textit{LO} WCET and a \textit{HI} WCET. A system can proceed at the \textit{LO} criticality level most of the time, assuming that tasks will exhibit their expected \textit{LO} WCETs. However, when any task exceeds its \textit{LO} WCET, the system enters the \textit{HI} criticality state and stops executing any \textit{LO} tasks. This means that to guarantee the safety of a system in the presence of unreliability, it is necessary to prove only that the \textit{HI} tasks will not exceed their \textit{HI} WCET times. The application adapts its behaviour based on the current state of the system by dropping less important activities and focusing on only the most important system guarantees. 

Prior work has applied mixed-criticality scheduling to wireless networks by assuming that it is the level of interference (i.e. the assumed maximum number of retransmissions required within a time period) that varies by criticality level~\cite{airtight}. This paper takes this idea and applies it to both the transmission of packets in a wireless communications swarm, but also to the behaviour of that swarm. Instead of treating criticality as just a feature of the scheduling and communications layer, this work argues that the application itself can respond to such changes usefully, based on developer-provided criticality requirements, in order to preserve the performance of the overall swarm. Such a characterisation can be exploited to allow for applications which assume communications will mostly operate well, and so behave accordingly, but can respond appropriately when, for whatever reason, communicating tasks start to fail to meet their deadlines. In Section~\ref{sec:application} we describe an example of such an application, with Section~\ref{sec:cohesion} describing how the application modifies its own behaviour based on feedback from the network layer.

\section{Prior Work}
\label{sec:priorwork}

In existing research, unrealistic assumptions are often made about the reliability of wireless communications, such as assuming perfect transmission within a given radius. Prior research has shown that imperfect communication results in swarm behaviours breaking down \cite{botnet}, and so it is therefore not always appropriate to simply ignore the communications medium and rely on TCP to transmit data `eventually'.

The MCS system model has been applied to industrial wireless communications~\cite{Changqing2017} to show that such an approach can give more reliable timing bounds by adapting the network routing based on the system's criticality level. This work provides better timing bounds, but is restricted to considering the communications layer only.

Our previous work has argued for the use of an MCS wireless protocol in swarm robotics applications \cite{circlepaper}. The system initially operates in \textit{LO} criticality mode, and if the level of successful message transmissions degrades past some threshold this is detected and the network switches to \textit{HI} criticality mode. The result of this change is typically to drop or deprioritise less important traffic. While this allows the application designer to preserve some behavioural elements in the presence of partial network degradation, it is inherently limited to preserving a subset of the full behaviour, and will eventually fail if network conditions continue to degrade. 

The AirTight protocol~\cite{airtight} considered in that work, being purely a point-to-point protocol, is impractical for real swarm robotics applications. We therefore introduce a model (Section \ref{sec:commodel}) and simulation implementation (Section \ref{sec:comprototype}) for a new protocol that can provide mixed criticality behaviour over broadcast transmissions.

The work in this paper extends this idea to consider the how the application's \emph{overall} behaviour can adapt to criticality. Specifically, we enable an application to observe the network layer's criticality level, so that the application can modify its own behaviour in response to current network conditions~(Section~\ref{sec:cohesion}).

\section{Communication Model}
\label{sec:commodel}

This paper argues for a closer integration of an application implementation with a mixed-criticality network MAC layer, specifically by allowing the application to be aware of the MAC layer's current criticality mode. This allows the application to detect when the network is no longer able to guarantee reliable message delivery due to communication faults, enabling it to adjust its behaviour to prevent further degradation. We compare the effect allowing a robot to use this criticality mode information to locally adjust its cohesion factor against simply using a static cohesion factor, as described in Section \ref{sec:cohesion}.

We consider a network of $N$ homogeneous robot nodes in which all transmissions are assumed to be broadcasts that should reliably reach each other node. The network provides multiple buffers of configurable priority and criticality such that the application designer can determine the order of transmissions. In each transmission slot, the node chooses to transmit the first message from the highest priority non-empty buffer at, or above, the current criticality level.

The network layer observes the number of ``failed'' transmissions, defined as transmissions after which it cannot prove that all nodes have received its last message. If the number of failed transmissions within a busy-period exceed a predefined threshold, the network switches to \textit{HI} criticality mode. Messages in \textit{LO} criticality buffers are discarded if the node enters \textit{HI} criticality mode. Time sensitive messages in \textit{HI} criticality buffers can optionally be assigned a maximum time-to-live (TTL), such that they expire and are not retransmitted after a given time period has elapsed since they were first queued.

\section{Communication Implementation}
\label{sec:comprototype}

In order to implement the behaviour described by the model in Section \ref{sec:commodel}, we introduce an implementation of AirTight~\cite{10.1145/3362987} modified to use broadcast transmission rather than point-to-point links. AirTight is a slot-based real-time, mixed criticality protocol which can guarantee the time-sensitive transmission of a set of traffic flows using ahead of time analysis under a given fault rate assumption.

The network runs using time-division multiplexing over 10ms time slots. There is a periodic slot table assigned a priori, such that the slot table is of length $N$ and each node has exactly one exclusive transmission slot. It is assumed that background clock synchronisation takes place such that nodes agree on the current slot.

Each node maintains a bit-field of length $N-1$, where each bit encodes whether the node successfully received a transmission in the last occurrence of the corresponding slot in the slot table (excluding the node's own transmission slot). This bit-field is included in the header of each transmission as a type of delayed acknowledgement, such that each receiving node can determine if its own last transmission was received by the sender. 

For each transmission buffer, the node maintains a further bit-field of length $N-1$ in which each bit encodes whether confirmation of successful delivery has been received from a corresponding other node. Upon reception of a frame, a node can thereby set the transmitting node's bit in its last transmission buffer's bitfield if the received header indicates that the transmission was received. Once all bits have been set, the message at the head of the buffer has been delivered to all other nodes. The message can then be removed from the buffer and the bit-field is cleared.

The node keeps two counters: a counter of the length of the current busy-period, and a counter of the number of retransmissions. The busy-period counter is incremented on all transmissions, whilst the retransmission counter is incremented whenever the node rebroadcasts a frame that had previously been transmitted. If the number of retransmissions exceeds a given threshold, the node switches to \textit{HI} criticality mode.

During its assigned transmission slot, a node broadcasts a single frame from the highest priority non-empty buffer at or above the current criticality level. If no such frame exists the counters are reset and, if the node was in \textit{HI} criticality mode, it switches back to \textit{LO}. The node then broadcasts a no-op frame such that the bit-field used for delayed acknowledgements is always transmitted. The real-time timing analysis of this implementation is future work but can be based on the structure of the original AirTight analysis.

The implementation of the simulation plugin, communications layer, and other artefacts associated with this paper can be found in our code repository \footnote{https://github.com/yorkrobotlab/argos3-airtight}.

\section{Exploration Application}
\label{sec:application}

We consider an autonomous wireless swarm robotics platform in which a swarm of robots should collaborate to explore an unknown area. The environment is partitioned into a two-dimensional grid. Robots should visit each cell in the grid and detect if there is an object present in that cell, building a map of clear/occupied cells in the area. It is assumed that obstacles are stationary, such that it is unimportant when a robots visits the cell, or whether a cell is visited multiple times by one or more robots. Each robot has its own local copy of the map, which it should complete for all accessible cells\footnote{The positioning of obstacles in the environment may render some cells inaccessible.}. Robots can communicate the sensed values for a given cell over the network, allowing other robots to insert this value into their map without needing to visit the cell.

For a concrete instantiation we consider a set of 10 Pi-Puck~\cite{pipuck} robots exploring a 6x6m grid of 10x10cm square cells. Pi-Puck robots only have simple infrared range-finder sensors that determine the distance to the closest object within a short range, but are unable to determine the nature of any detected object. Therefore, to avoid falsely detecting other robots as an obstacle in the environment, robots must maintain a minimum separation. Each robot is assumed to be able to determine its own location, which it must communicate to the other robots to avoid such near-collisions. Robots cannot store their complete location history, so they cannot retrospectively determine that incorrect data may have been sensed where position messages have been lost or delayed. These position messages are therefore intuitively subject to soft real-time constraints, since robots must learn the positions of other robots before the minimum separation distance is violated.

The robot behaviour is loosely inspired by an existing algorithm~\cite{tran2022}, but adapted to the much simpler Pi-Puck robots and to much reduced communication ability. It is implemented by picking and driving towards a target cell, which is always chosen as one of the nine cells within a 3x3 grid centred on the robot's current position. A new target cell is chosen once the current target is reached, or the robot encounters an obstacle such that it deems the current target to be unreachable. The target cell is selected as the cell for which the sum of the following weights results in the smallest value. 

\begin{itemize}
    \item Diagonal movements are assigned a weight of +1.
    \item The cell the robot is currently in is assigned a linearly increasing weight the longer it remains in that cell, and the cell it was in immediately previously is assigned a weight of +1.
    \item An avoidance score of +1000 if the cell is known to contain an obstacle.
    \item An attraction score of -10 if it is an unexplored cell.
    \item A separation score of 400000, 200000, 100000, 4, 1, 0.25, or 0.1 respectively if the distance to the closest target cell of another robot, counted as a number of cells, between 0 and 6. 
    \item An alignment score, given by the dot product of the robot's forwards vector and the vector from it's current position to the potential target.
    \item An attraction score equal to the distance to the closest reachable unexplored cell, counted as a number of cells.
    \item Optionally, a cohesion force of $8d^3$, where $d$ is the distance to the computed centroid of all robots using the most recent position information the robot has received. This force is applied according to the rules defined in Section \ref{sec:cohesion}.
\end{itemize}

\newpage
\subsection{Robot Cohesion}
\label{sec:cohesion}

\begin{figure}[h]
\centering
\includegraphics[width=0.9\textwidth]{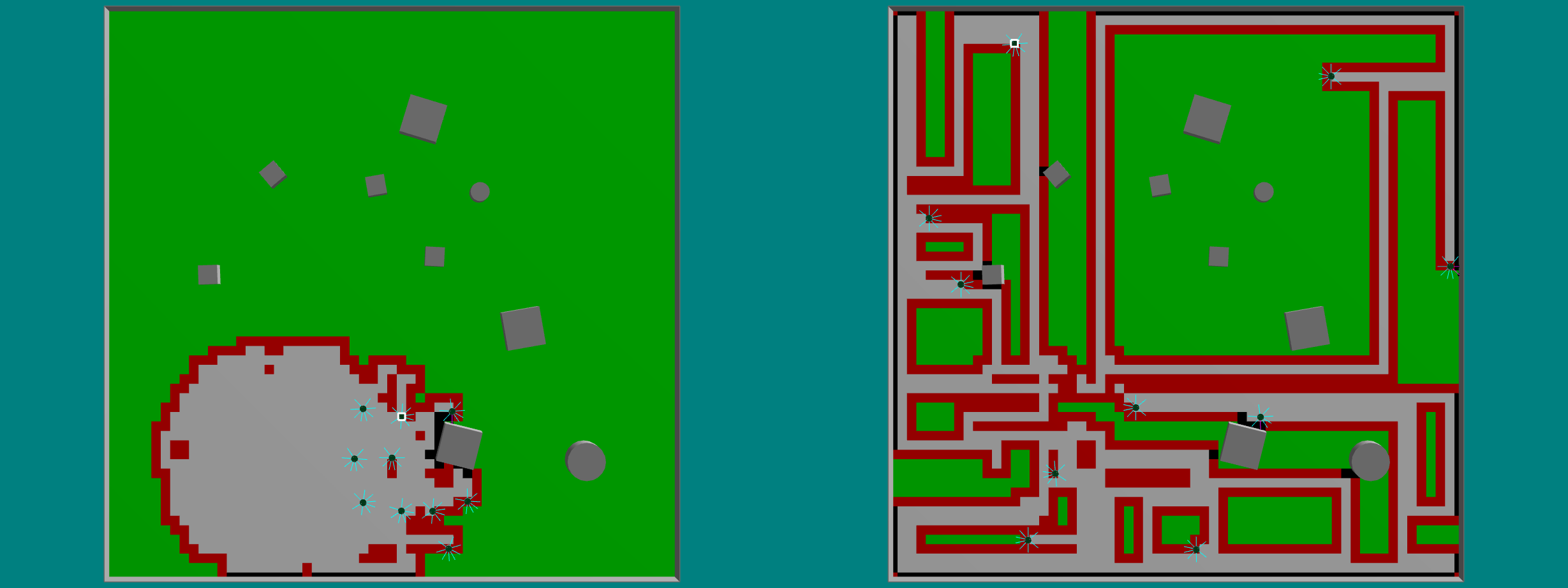}
\caption{Visualisation of exploration state after 95s when applying constant cohesion (left) compared with no cohesion (right) under perfect network conditions. Explored cells are shown in grey if empty or black if containing/bordering an obstacle; unexplored cells are shown in red when bordering explored cells, else in green.}
\label{fig:cohesion}
\end{figure}

The effect of adding a cohesion element to the target cell selection depends on the environmental factors encountered by the robots. Under perfect communication, robot cohesion reduces the overall application performance. This can be intuitively understood by considering that robots that disperse maximally will not block each other and can greedily explore new cells. The closer the robots are pulled together, the more often robots may need to change direction (for example by revisiting already explored cells) in order to avoid violating the minimum separation constraint. As shown in Figure \ref{fig:cohesion}, this may result in robots towards the rear of a group being surrounded by already explored cells.

With a communication model that deteriorates with distance, the performance of a solution where robots disperse decreases, since robots are unable to receive the sensing information of other robots. In extreme cases, this could effectively result in each robot needing to explore the entire area. There is therefore a trade-off in which some amount cohesion is useful to preserve communication, but too much cohesion decreases performance by limiting the exploration ability. The optimal level of cohesion depends on the properties of the wireless medium, which in a real scenario may not be known a priori. 

In this paper, we argue for application adaptation based on the state of the network, by applying cohesion weighting on target cell selection using a mixed-criticality approach. While the network layer is in \textit{LO} criticality mode the robots can disperse to maximise their exploration potential, before being pulled back towards each other if/when the network changes to \textit{HI} criticality mode. Once the network has recovered, the robots can then resume exploring. This results in an equilibrium that allows the robots to adapt their behaviour to the encountered conditions. We compare the effects of the following four ways of applying the cohesion weighting:
\begin{itemize}
  \item No cohesion: The cohesion weight is completely disabled.
  \item Constant cohesion: A cohesion weight is always applied.
  \item Half cohesion: The cohesion weight is always applied but is computed using half of the true centroid distance.
  \item Mixed criticality: The application applies a cohesion weighting when the network protocol has been in \textit{HI} criticality mode at any point within the last three seconds.
\end{itemize}

\section{Simulation Setup}

\subsection{Simulation Framework}

Our simulation setup is based on the popular ARGoS robot simulator~\cite{argos}. An extended version of a custom networking plugin \cite{circlepaper} implements the network simulation capabilities. Each simulation step is assumed to correspond to a network level transmission slot in which a robot can attempt to send or receive a single message. A simulated transmission model controls which messages are successfully received.

\begin{figure}
\centering
\begin{minipage}{0.38\textwidth}
\centering
\includegraphics[width=\textwidth]{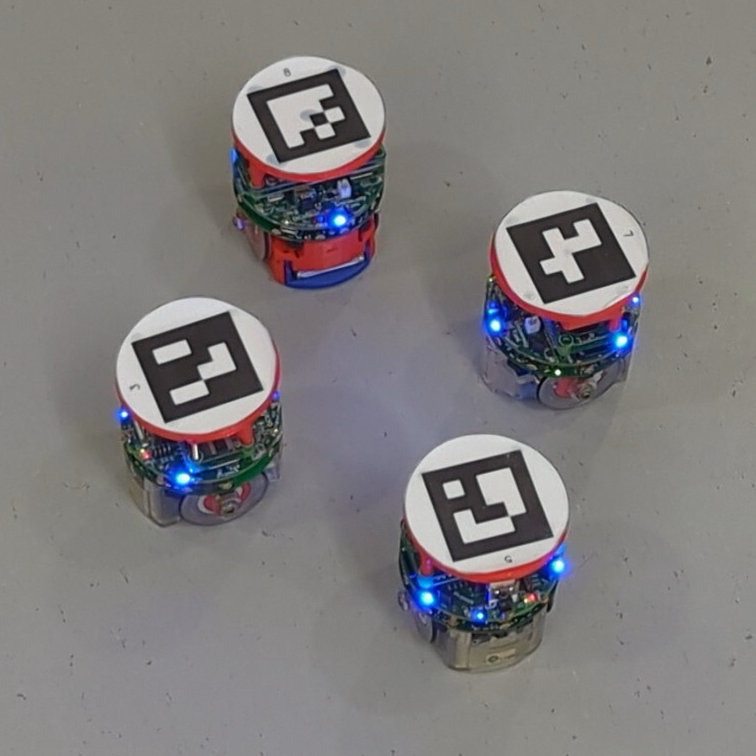}
\end{minipage}
\begin{minipage}{0.38\textwidth}
\centering
\includegraphics[width=\textwidth]{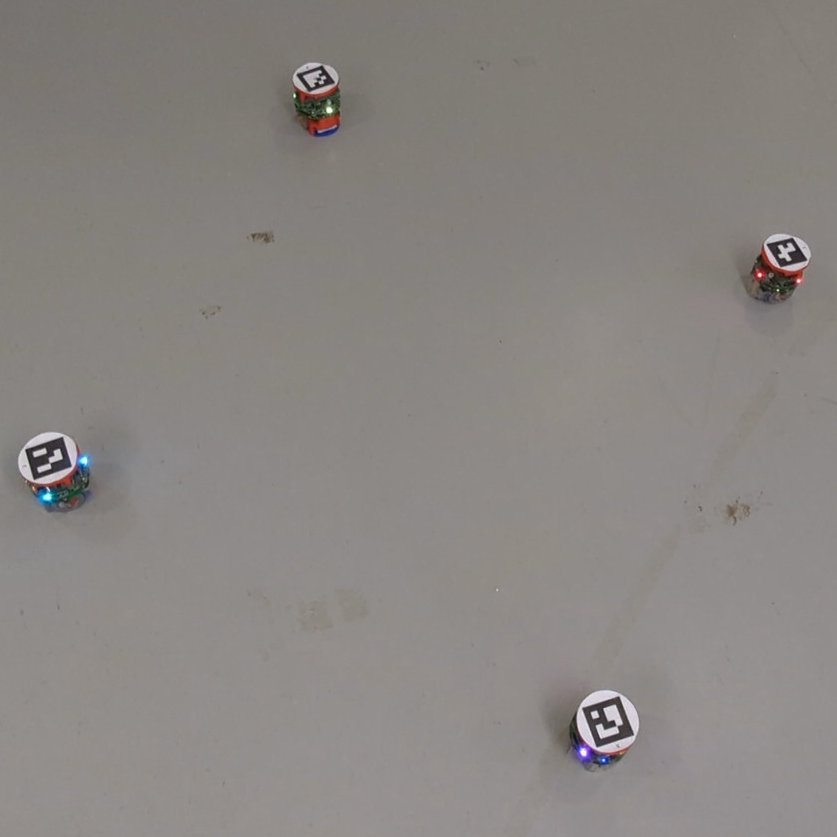}
\end{minipage}
\caption{The running ``physicalised simulation'' of an earlier experiment on the Pi-Puck robots. The simulation shows the LED synchronisation breaking down as the robot separation increases.}
\label{fig:pipucks}
\end{figure}

In order to partially validate our simulation framework, we have implemented a ``physicalised simulation'' of our earlier experiments using real Pi-Puck robots (Figure~\ref{fig:pipucks}). By this we mean an implementation on physical robots, but where some key components are still simulated. Specifically, we use Pi-Puck robots communicating over IEEE 802.15.4 provided by XBee modules, whilst simulating the packet loss by artifically discarding some messages according to the packet delivery rates provided by the transmission model. Since this still relies on a simulated transmission model, the results between the full simulation and physicalised simulation are very similar. Due to the logistical challenges of running larger scale experiments with physical robots, we have not yet implemented such a physicalised simulation for the current experiments. 

\subsection{Robot Configuration}

Each robot is configured with two network buffers, for position messages and cell status message. A position message contains the robots target location, while a cell status message encodes a single cell to be either clear or containing an obstacle. Since the focus of these experiments is on the effect of application level behaviour changes based on the network criticality level, we configure both buffers as \textit{HI} criticality to prevent criticality changes having an impact at the network level. Positional data is time sensitive to ensure the minimum robot separation is preserved, and so this buffer is configured with the higher priority. Positional messages are set to be retried for a maximum of 0.8s to prevent old position data filling up transmission buffers, whilst cell status messages are set not to expire since these are not time sensitive.

\subsection{Arena Generation}

We randomly generate 100 simulated arenas, comparing the exploration performance of each cohesion and transmission model configuration combination across these arenas. The robots are placed first, by distributing them within a 1.8m by 1.8m square centred around a random point in the area, whilst ensuring each robot placed at least 30cm away from the next closest robot. We then distribute up to 17 obstacles across the arena, again requiring a minimum separation from each other obstacle and each robot starting position.

\subsection{Transmission Model}

Packet delivery is determined by a simulated transmission model in which successful or unsuccessful delivery is determined independently for each transmission and each potential receiving node. The packet delivery rate is assumed to be fixed to a maximum of 95\% for distance of less than 0.5m, after which the packet delivery rate is inversely proportional to the square distance between the nodes, subject to an additional constant scaling factor ($k$). By modifying this scaling factor, the effect of the packet delivery rate on the application behaviour can be observed.

\begin{figure}[htbp]
\centering
\begin{minipage}{0.4\textwidth}
\[ \text{PDR} = \frac{0.95}{1 + ( k\cdot X )^2} \]
\hspace*{1cm}Where:
\[ X = \begin{cases}
0 & d \leq 0.5 \\
d - 0.5 & d > 0.5
\end{cases} \]
\end{minipage}%
\begin{minipage}{0.6\textwidth}
\centering
\includegraphics[width=\textwidth]{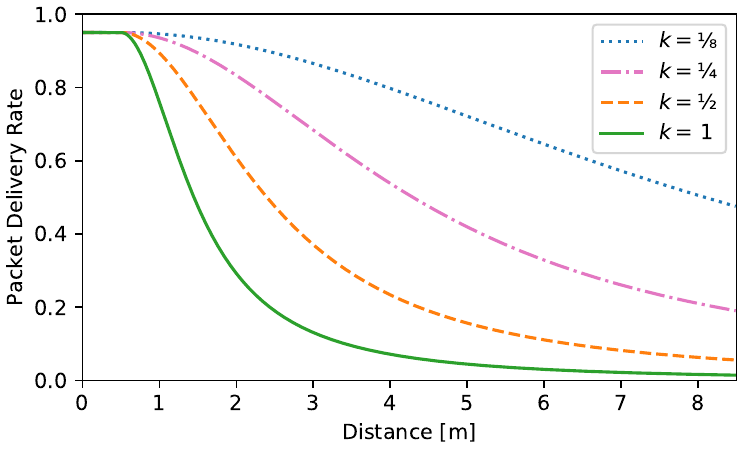}
\end{minipage}
\caption{Transmission model defining packet reception probability in relation to distance, for scaling factors $k=\frac{1}{8}$, $k=\frac{1}{4}$, $k=\frac{1}{2}$, and $k=1$. }
\label{fig:transmodel}
\end{figure}

This is a simple model that cannot capture the true complexity of real-world wireless communications. Prior research \cite{radiale, scale} has shown that observed packet delivery rates do not correlate as strongly with distance in real-world experiments. Nonetheless, observed results broadly show that there exists a safe distance cutoff up to which the communication is generally reliable. Baccour et al. describes this as the ``connected'' region~\cite{radiale}, which is followed by ``transitional'' and ``disconnected'' regions where packet delivery first becomes intermittent and then mostly unsuccessful. This fits well with the mixed criticality system model, which inherently assumes that the system will operate under its optimistic assumptions most of the time, before encountering some kind of state change that requires rectification.

The implementation here does not rely on specific details of the transmission model, and the criticality response merely assumes that moving closer to other nodes is likely to improve packet reception rates. We therefore believe the simulation results should be broadly applicable regardless of the selected transmission model.

\section{Results}

\begin{figure}[htb]
\centering
\begin{minipage}{.48\textwidth}
  \centering
  \includegraphics[width=\textwidth]{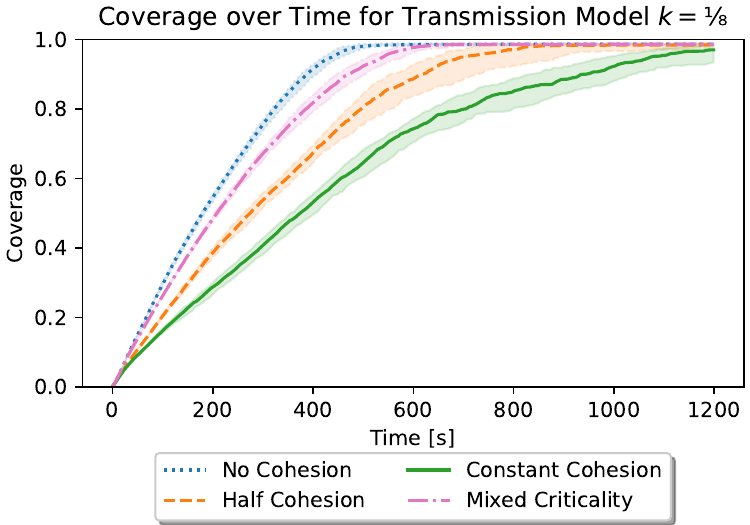}
\end{minipage}%
\begin{minipage}{.48\textwidth}
  \centering
  \includegraphics[width=\textwidth]{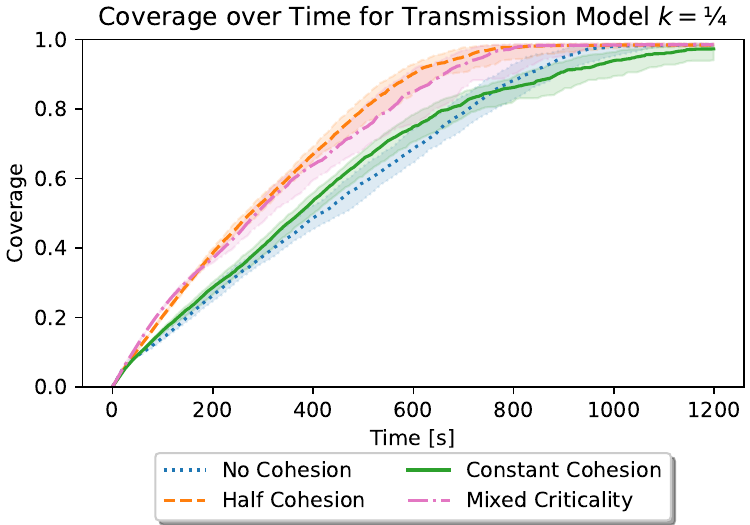}
\end{minipage}
\caption{Coverage over time across the four cohesion modes for transmission model scaling factors $k=\frac{1}{8}$ and $k=\frac{1}{4}$. Line shows median value; shaded region shows interquartile range across 100 simulation runs.}
\label{fig:dropoff8and4}
\end{figure}

With the transmission model scaling factor configured for the most gradual dropoff in packet delivery rates, $k=\frac{1}{8}$, the simulation results in Figure \ref{fig:dropoff8and4} shows that not applying a cohesion force results in the highest performance. At this low dropoff rate the robots can adequately communicate across the entire arena such that there is no advantage to moving as a cohesive group. The mixed criticality configuration can spend a significant proportion of its runtime in \textit{LO} criticality mode (where no cohesion is applied) and thus displays better performance than the half cohesion or constant cohesion configurations. 

When the transmission model scaling factor is increased to $k=\frac{1}{4}$ a maximal dispersion of the nodes begins to impede communication between the nodes, creating an advantage to applying some level of cohesion. The mixed criticality and half cohesion configurations perform similarly to each other. The no cohesion configuration starts to suffer from the aforementioned communication issues, while the performance of the constant cohesion configuration is still reduced from applying a stronger cohesive force than necessary.

\begin{figure}[htb]
\centering
\begin{minipage}{.48\textwidth}
  \centering
  \includegraphics[width=\textwidth]{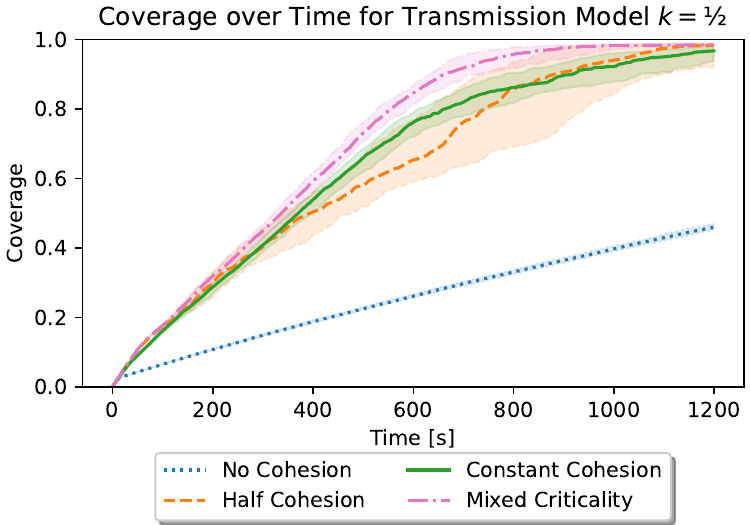}
\end{minipage}%
\begin{minipage}{.48\textwidth}
  \centering
  \includegraphics[width=\textwidth]{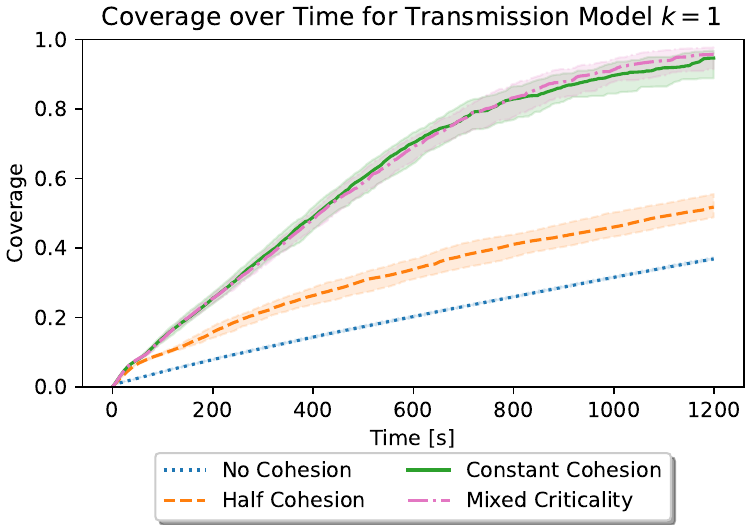}
\end{minipage}
\caption{Coverage over time across the four cohesion modes for transmission model scaling factors $k=\frac{1}{2}$ and $k=1$. Line shows median value; shaded region shows interquartile range across 100 simulation runs.}
\label{fig:dropoff2and1}
\end{figure}

A further increase of the transmission model scaling factor to $k=\frac{1}{2}$, shown in Figure \ref{fig:dropoff2and1}, causes the no cohesion configuration to break down. The performance of the half cohesion configuration is also decreased, showing a fairly wide spread depending on runtime circumstances. The mixed criticality configuration is able to adapt to the conditions, yielding the highest performance for this transmission model.

At the maximum tested dropoff rate, with the transmission model scaling factor set to $k=1$, the mixed criticality mode and constant cohesion modes perform very similarly, with the low packet delivery rates causing the mixed criticality configuration to spend large proportions of its time in \textit{HI} criticality mode where cohesion is applied. The half cohesion and no cohesion modes both show poor performance, with the half cohesion mode no longer resulting in a sufficiently tight formation to maintain communication between the nodes.

The constant cohesion configuration is mostly unaffected by the transmission model's scaling factor. While this results in comparatively good performance when communications are limited to short ranges, it cannot take advantage of long communication ranges under good radio conditions. The half cohesion configuration provides comparatively good performance for the two middling transmission models, but neither takes full advantage of long communication ranges nor does it cope with very short communication ranges.

The mixed criticality mode is better able to adjust to different communication dropoff rates, and can do so in a way that does not require the application layer to understand the specific issues that are happening at the networking layer. While a static cohesion factor can provide similar levels of performance for any given transmission model, this requires the transmission characteristics to be known beforehand and to remain static. For a real-world application this is unlikely to be the case. A static cohesion factor therefore requires the application designer to make a tradeoff between performance under good networking conditions and reliability under network degradation. A mixed criticality approach avoids this tradeoff by allowing the application to observe the true conditions at runtime.

\section{Limitations and Future Work}

The communication protocol presented in this paper is developed to study the effect of adjusting application behaviour based on a network layer criticality level. Compared to a complete protocol it is lacking in several aspects, most importantly the absence of formal timing analysis that can provide guarantees on the network performance. In future work we intend to develop a real-time mixed-criticality protocol that supports network layer and application integration while being suitable for swarm robotics and providing such timing guarantees.

We also intend to further study the two-way relationship between application behaviour and network MAC layer. In the implementation provided in this paper, the application is aware of the current network layer criticality level, and has been programmed that cohesion should be applied only when the network layer is in \textit{HI} criticality mode. Beyond this simple rule, however, it could be imagined a more intelligent application could attempt to predict the effect of its future behaviour on the network. This could allow the application to either modify its planned behaviour to avoid network issues, or warn the network layer such that it could prepare for a drop in packet delivery rates.

\section{Conclusion}

In this paper we have presented a mixed-criticality approach to swarm robotics application behaviour. Mixed criticality is widely applied in real-time CPU scheduling and network protocols, but has so far not been widely applied to swarm robotics. The parameters controlling a robot's behaviour may have different optimal values depending on the conditions encountered by the robot. Communication conditions at runtime can vary from those expected by the application designer in ways that are often opaque to the application, placing stress on the network and leading to a loss of real-time performance. A mixed-criticality approach allows a systematic way for the application to both define what is considered of higher and lower importance, and then respond in a way that prioritises resources appropriately.

Our simulation results show that a swarm robotics application using a mixed-criticality approach to adjust its behaviour to match the encountered conditions can be made more robust than one that uses a static configuration. In future work we intend to develop mixed-criticality network protocols targeted at swarm robotics applications and further study the integration of network and application behaviour.

\bibliographystyle{eptcs}
\bibliography{refs}

\begin{thebibliography}{10}
\providecommand{\bibitemdeclare}[2]{}
\providecommand{\surnamestart}{}
\providecommand{\surnameend}{}
\providecommand{\urlprefix}{Available at }
\providecommand{\url}[1]{\texttt{#1}}
\providecommand{\href}[2]{\texttt{#2}}
\providecommand{\urlalt}[2]{\href{#1}{#2}}
\providecommand{\doi}[1]{doi:\urlalt{https://doi.org/#1}{#1}}
\providecommand{\eprint}[1]{arXiv:\urlalt{https://arxiv.org/abs/#1}{#1}}
\providecommand{\bibinfo}[2]{#2}

\bibitemdeclare{article}{radiale}
\bibitem{radiale}
\bibinfo{author}{Nouha \surnamestart Baccour\surnameend}, \bibinfo{author}{Anis
  \surnamestart Koubâa\surnameend}, \bibinfo{author}{Maissa \surnamestart {Ben
  Jamâa}\surnameend}, \bibinfo{author}{Denis \surnamestart {do
  Rosário}\surnameend}, \bibinfo{author}{Habib \surnamestart
  Youssef\surnameend}, \bibinfo{author}{Mário \surnamestart Alves\surnameend}
  \& \bibinfo{author}{Leandro~B. \surnamestart Becker\surnameend}
  (\bibinfo{year}{2011}): \emph{\bibinfo{title}{{RadiaLE}: A framework for
  designing and assessing link quality estimators in wireless sensor
  networks}}.
\newblock {\slshape \bibinfo{journal}{Ad Hoc Networks}}
  \bibinfo{volume}{9}(\bibinfo{number}{7}), pp. \bibinfo{pages}{1165--1185},
  \doi{10.1016/j.adhoc.2011.01.006}.

\bibitemdeclare{inproceedings}{bjerknes2013fault}
\bibitem{bjerknes2013fault}
\bibinfo{author}{Jan~Dyre \surnamestart Bjerknes\surnameend} \&
  \bibinfo{author}{Alan~FT \surnamestart Winfield\surnameend}
  (\bibinfo{year}{2013}): \emph{\bibinfo{title}{On fault tolerance and
  scalability of swarm robotic systems}}.
\newblock In: {\slshape \bibinfo{booktitle}{Distributed Autonomous Robotic
  Systems: The 10th International Symposium}},
  \bibinfo{organization}{Springer}, pp. \bibinfo{pages}{431--444},
  \doi{10.1007/978-3-642-32723-0_31}.

\bibitemdeclare{inproceedings}{airtight}
\bibitem{airtight}
\bibinfo{author}{Alan \surnamestart Burns\surnameend}, \bibinfo{author}{James
  \surnamestart Harbin\surnameend}, \bibinfo{author}{Leandro \surnamestart
  Indrusiak\surnameend}, \bibinfo{author}{Iain \surnamestart Bate\surnameend},
  \bibinfo{author}{Rob \surnamestart Davis\surnameend} \&
  \bibinfo{author}{David \surnamestart Griffin\surnameend}
  (\bibinfo{year}{2018}): \emph{\bibinfo{title}{AirTight: A Resilient Wireless
  Communication Protocol for Mixed-Criticality Systems}}.
\newblock In: {\slshape \bibinfo{booktitle}{2018 IEEE 24th International
  Conference on Embedded and Real-Time Computing Systems and Applications
  (RTCSA)}}, pp. \bibinfo{pages}{65--75}, \doi{10.1109/RTCSA.2018.00017}.

\bibitemdeclare{techreport}{scale}
\bibitem{scale}
\bibinfo{author}{Alberto \surnamestart Cerpa\surnameend}, \bibinfo{author}{Naim
  \surnamestart Busek\surnameend} \& \bibinfo{author}{Deborah \surnamestart
  Estrin\surnameend} (\bibinfo{year}{2003}): \emph{\bibinfo{title}{SCALE: A
  tool for Simple Connectivity Assessment in Lossy Environments}}.
\newblock \bibinfo{type}{Technical Report}.
\newblock \urlprefix\url{https://escholarship.org/uc/item/2g49z78g}.

\bibitemdeclare{article}{international2016industrial}
\bibitem{international2016industrial}
\bibinfo{author}{International~Electrotechnical \surnamestart
  Commission\surnameend} et~al. (\bibinfo{year}{2016}):
  \emph{\bibinfo{title}{Industrial Networks—Wireless Communication Network
  and Communication Profiles—WirelessHART (IEC 62591: 2016)}}.
\newblock {\slshape \bibinfo{journal}{IEC: Geneva, Switzerland}}
  \bibinfo{volume}{3}, pp. \bibinfo{pages}{1--1043}.

\bibitemdeclare{inproceedings}{10.1145/2426656.2426658}
\bibitem{10.1145/2426656.2426658}
\bibinfo{author}{Federico \surnamestart Ferrari\surnameend},
  \bibinfo{author}{Marco \surnamestart Zimmerling\surnameend},
  \bibinfo{author}{Luca \surnamestart Mottola\surnameend} \&
  \bibinfo{author}{Lothar \surnamestart Thiele\surnameend}
  (\bibinfo{year}{2012}): \emph{\bibinfo{title}{Low-Power Wireless Bus}}.
\newblock In: {\slshape \bibinfo{booktitle}{Proceedings of the 10th ACM
  Conference on Embedded Network Sensor Systems}}, \bibinfo{series}{SenSys
  '12}, \bibinfo{publisher}{Association for Computing Machinery},
  \bibinfo{address}{New York, NY, USA}, p. \bibinfo{pages}{1–14},
  \doi{10.1145/2426656.2426658}.

\bibitemdeclare{inproceedings}{5779066}
\bibitem{5779066}
\bibinfo{author}{Federico \surnamestart Ferrari\surnameend},
  \bibinfo{author}{Marco \surnamestart Zimmerling\surnameend},
  \bibinfo{author}{Lothar \surnamestart Thiele\surnameend} \&
  \bibinfo{author}{Olga \surnamestart Saukh\surnameend} (\bibinfo{year}{2011}):
  \emph{\bibinfo{title}{Efficient network flooding and time synchronization
  with Glossy}}.
\newblock In: {\slshape \bibinfo{booktitle}{Proceedings of the 10th ACM/IEEE
  International Conference on Information Processing in Sensor Networks}}, pp.
  \bibinfo{pages}{73--84}.

\bibitemdeclare{article}{10.1145/3362987}
\bibitem{10.1145/3362987}
\bibinfo{author}{J.~\surnamestart Harbin\surnameend},
  \bibinfo{author}{A.~\surnamestart Burns\surnameend}, \bibinfo{author}{R.~I.
  \surnamestart Davis\surnameend}, \bibinfo{author}{L.~S. \surnamestart
  Indrusiak\surnameend}, \bibinfo{author}{I.~\surnamestart Bate\surnameend} \&
  \bibinfo{author}{D.~\surnamestart Griffin\surnameend} (\bibinfo{year}{2019}):
  \emph{\bibinfo{title}{The AirTight Protocol for Mixed Criticality Wireless
  CPS}}.
\newblock {\slshape \bibinfo{journal}{ACM Trans. Cyber-Phys. Syst.}}
  \bibinfo{volume}{4}(\bibinfo{number}{2}), \doi{10.1145/3362987}.

\bibitemdeclare{inproceedings}{pipuck}
\bibitem{pipuck}
\bibinfo{author}{Alan~G. \surnamestart Millard\surnameend},
  \bibinfo{author}{Russell \surnamestart Joyce\surnameend},
  \bibinfo{author}{James~A. \surnamestart Hilder\surnameend},
  \bibinfo{author}{Cristian \surnamestart Fleşeriu\surnameend},
  \bibinfo{author}{Leonard \surnamestart Newbrook\surnameend},
  \bibinfo{author}{Wei \surnamestart Li\surnameend}, \bibinfo{author}{Liam~J.
  \surnamestart McDaid\surnameend} \& \bibinfo{author}{David~M. \surnamestart
  Halliday\surnameend} (\bibinfo{year}{2017}): \emph{\bibinfo{title}{The
  Pi-puck extension board: A raspberry Pi interface for the e-puck robot
  platform}}.
\newblock In: {\slshape \bibinfo{booktitle}{2017 IEEE/RSJ International
  Conference on Intelligent Robots and Systems (IROS)}}, pp.
  \bibinfo{pages}{741--748}, \doi{10.1109/IROS.2017.8202233}.

\bibitemdeclare{inproceedings}{pinciroli2016buzz}
\bibitem{pinciroli2016buzz}
\bibinfo{author}{Carlo \surnamestart Pinciroli\surnameend} \&
  \bibinfo{author}{Giovanni \surnamestart Beltrame\surnameend}
  (\bibinfo{year}{2016}): \emph{\bibinfo{title}{Buzz: An extensible programming
  language for heterogeneous swarm robotics}}.
\newblock In: {\slshape \bibinfo{booktitle}{2016 IEEE/RSJ International
  Conference on Intelligent Robots and Systems (IROS)}},
  \bibinfo{organization}{IEEE}, pp. \bibinfo{pages}{3794--3800},
  \doi{10.1109/IROS.2016.7759558}.

\bibitemdeclare{article}{argos}
\bibitem{argos}
\bibinfo{author}{Carlo \surnamestart Pinciroli\surnameend},
  \bibinfo{author}{Vito \surnamestart Trianni\surnameend},
  \bibinfo{author}{Rehan \surnamestart O'Grady\surnameend},
  \bibinfo{author}{Giovanni \surnamestart Pini\surnameend},
  \bibinfo{author}{Arne \surnamestart Brutschy\surnameend},
  \bibinfo{author}{Manuele \surnamestart Brambilla\surnameend},
  \bibinfo{author}{Nithin \surnamestart Mathews\surnameend},
  \bibinfo{author}{Eliseo \surnamestart Ferrante\surnameend},
  \bibinfo{author}{Gianni \surnamestart {Di Caro}\surnameend},
  \bibinfo{author}{Frederick \surnamestart Ducatelle\surnameend},
  \bibinfo{author}{Mauro \surnamestart Birattari\surnameend},
  \bibinfo{author}{Luca~Maria \surnamestart Gambardella\surnameend} \&
  \bibinfo{author}{Marco \surnamestart Dorigo\surnameend}
  (\bibinfo{year}{2012}): \emph{\bibinfo{title}{{ARGoS}: a Modular, Parallel,
  Multi-Engine Simulator for Multi-Robot Systems}}.
\newblock {\slshape \bibinfo{journal}{Swarm Intelligence}}
  \bibinfo{volume}{6}(\bibinfo{number}{4}), pp. \bibinfo{pages}{271--295},
  \doi{10.1007/s11721-012-0072-5}.

\bibitemdeclare{inproceedings}{botnet}
\bibitem{botnet}
\bibinfo{author}{Mark \surnamestart Selden\surnameend}, \bibinfo{author}{Jason
  \surnamestart Zhou\surnameend}, \bibinfo{author}{Felipe \surnamestart
  Campos\surnameend}, \bibinfo{author}{Nathan \surnamestart
  Lambert\surnameend}, \bibinfo{author}{Daniel \surnamestart Drew\surnameend}
  \& \bibinfo{author}{Kristofer S.~J. \surnamestart Pister\surnameend}
  (\bibinfo{year}{2021}): \emph{\bibinfo{title}{BotNet: A Simulator for
  Studying the Effects of Accurate Communication Models on Multi-Agent and
  Swarm Control}}.
\newblock In: {\slshape \bibinfo{booktitle}{2021 International Symposium on
  Multi-Robot and Multi-Agent Systems (MRS)}}, pp. \bibinfo{pages}{101--109},
  \doi{10.1109/MRS50823.2021.9620611}.

\bibitemdeclare{inproceedings}{circlepaper}
\bibitem{circlepaper}
\bibinfo{author}{Sven \surnamestart Signer\surnameend},
  \bibinfo{author}{Alan~G. \surnamestart Millard\surnameend} \&
  \bibinfo{author}{Ian \surnamestart Gray\surnameend} (\bibinfo{year}{2022}):
  \emph{\bibinfo{title}{Mixed-Criticality Wireless Communication for Robot
  Swarms}}.
\newblock In: {\slshape \bibinfo{booktitle}{Proceedings of the 9th
  International Workshop on Mixed Criticality Systems}}, pp.
  \bibinfo{pages}{20--25}.
\newblock
  \urlprefix\url{https://wmc2022.github.io/assets/WMC_2022_Proceedings.pdf}.

\bibitemdeclare{article}{tran2022}
\bibitem{tran2022}
\bibinfo{author}{Vu~Phi \surnamestart Tran\surnameend},
  \bibinfo{author}{Matthew~A. \surnamestart Garratt\surnameend},
  \bibinfo{author}{Kathryn \surnamestart Kasmarik\surnameend},
  \bibinfo{author}{Sreenatha~G. \surnamestart Anavatti\surnameend} \&
  \bibinfo{author}{Shadi \surnamestart Abpeikar\surnameend}
  (\bibinfo{year}{2022}): \emph{\bibinfo{title}{Frontier-led swarming: Robust
  multi-robot coverage of unknown environments}}.
\newblock {\slshape \bibinfo{journal}{Swarm and Evolutionary Computation}}
  \bibinfo{volume}{75}, p. \bibinfo{pages}{101171},
  \doi{10.1016/j.swevo.2022.101171}.

\bibitemdeclare{inproceedings}{Vestal2007}
\bibitem{Vestal2007}
\bibinfo{author}{Steve \surnamestart Vestal\surnameend} (\bibinfo{year}{2007}):
  \emph{\bibinfo{title}{Preemptive Scheduling of Multi-criticality Systems with
  Varying Degrees of Execution Time Assurance}}.
\newblock In: {\slshape \bibinfo{booktitle}{28th IEEE International Real-Time
  Systems Symposium (RTSS 2007)}}, pp. \bibinfo{pages}{239--243},
  \doi{10.1109/RTSS.2007.47}.

\bibitemdeclare{article}{Changqing2017}
\bibitem{Changqing2017}
\bibinfo{author}{Changqing \surnamestart Xia\surnameend},
  \bibinfo{author}{Xi~\surnamestart Jin\surnameend}, \bibinfo{author}{Linghe
  \surnamestart Kong\surnameend} \& \bibinfo{author}{Peng \surnamestart
  Zeng\surnameend} (\bibinfo{year}{2017}): \emph{\bibinfo{title}{Bounding the
  Demand of Mixed-Criticality Industrial Wireless Sensor Networks}}.
\newblock {\slshape \bibinfo{journal}{IEEE Access}} \bibinfo{volume}{5}, pp.
  \bibinfo{pages}{7505--7516}, \doi{10.1109/ACCESS.2017.2654483}.

\bibitemdeclare{article}{10.1145/3012005}
\bibitem{10.1145/3012005}
\bibinfo{author}{Marco \surnamestart Zimmerling\surnameend},
  \bibinfo{author}{Luca \surnamestart Mottola\surnameend},
  \bibinfo{author}{Pratyush \surnamestart Kumar\surnameend},
  \bibinfo{author}{Federico \surnamestart Ferrari\surnameend} \&
  \bibinfo{author}{Lothar \surnamestart Thiele\surnameend}
  (\bibinfo{year}{2017}): \emph{\bibinfo{title}{Adaptive Real-Time
  Communication for Wireless Cyber-Physical Systems}}.
\newblock {\slshape \bibinfo{journal}{ACM Trans. Cyber-Phys. Syst.}}
  \bibinfo{volume}{1}(\bibinfo{number}{2}), \doi{10.1145/3012005}.

\end{thebibliography}
\end{document}